\documentclass[sigconf]{acmart}

\usepackage{graphicx}
\graphicspath{{img/}}
\usepackage{subcaption}
\usepackage{url}
\usepackage{amsmath}
\usepackage{stfloats}

\usepackage{acronym}

\newacro{SNN}{Spiking Neural Network}
\newacro{ANN}{Artificial Neural Network}
\newacro{STDP}{Spike Timing Dependent Plasticity}
\newacro{PE}{Processing Element}
\newacro{PS}{Procesing System}
\newacro{PL}{Programmable Logic}
\newacro{PE}{Processing Element}
\newacro{FPGA}{Field Programmable Gate Array}
\newacro{IF}{Integrate and Fire}
\newacro{LTP}{Long Term Potentiation}
\newacro{LTD}{Long Term Depression}
\newacro{DoG}{Difference of Gaussians}
\newacro{TTFS}{Time To First Spike}
\newacro{CNN}{Convolutional Neural Networks}
\newacro{LTPS}{Learning Time Per Sample}
\newacro{SVM}{Support Vector Machine}

\setcopyright{rightsretained} 

\AtBeginDocument{%
	\providecommand\BibTeX{{%
			\normalfont B\kern-0.5em{\scshape i\kern-0.25em b}\kern-0.8em\TeX}}}

\begin{document}

\title{Connection Pruning for Deep Spiking Neural Networks with On-Chip Learning}

\author{Thao N.N. Nguyen}
\affiliation{%
	\institution{National University of Singapore}
	\streetaddress{4 Engineering Drive 3}
	\country{Singapore}
	\postcode{S117583}
}
\email{thaonnnguyen@u.nus.edu}
\author{Bharadwaj Veeravalli}
\affiliation{%
	\institution{National University of Singapore}
	\streetaddress{4 Engineering Drive 3}
	\country{Singapore}
	\postcode{S117583}
}
\email{elebv@nus.edu.sg}
\author{Xuanyao Fong}
\affiliation{%
	\institution{National University of Singapore}
	\streetaddress{4 Engineering Drive 3}
	\country{Singapore}
	\postcode{S117583}
}
\email{kelvin.xy.fong@nus.edu.sg}

\renewcommand{\shortauthors}{Nguyen et al.}

\begin{abstract}
Long training time hinders the potential of the deep, large-scale \ac{SNN} with the on-chip learning capability to be realized on the embedded systems hardware. Our work proposes a novel connection pruning approach that can be applied during the on-chip \ac{STDP}-based learning to optimize the learning time and the network connectivity of the deep \ac{SNN}. We applied our approach to a deep \ac{SNN} with the \ac{TTFS} coding and has successfully achieved 2.1x speed-up and 64\% energy savings in the on-chip learning and reduced the network connectivity by 92.83\%, without incurring any accuracy loss. Moreover, the connectivity reduction results in 2.83x speed-up and 78.24\% energy savings in the inference. Evaluation of our proposed approach on the \ac{FPGA} platform revealed 0.56\% power overhead was needed to implement the pruning algorithm.
\end{abstract}

\copyrightyear{2021}
\acmYear{2021}
\acmConference[ICONS 2021]{International Conference on Neuromorphic Systems 2021}{July 27--29, 2021}{Knoxville, TN, USA}
\acmBooktitle{International Conference on Neuromorphic Systems 2021 (ICONS 2021), July 27--29, 2021, Knoxville, TN, USA}
\acmDOI{10.1145/3477145.3477157}
\acmISBN{978-1-4503-8691-3/21/07}


\maketitle
\pagestyle{plain}

\section{Introduction}
Spiking Neural Network (SNN) has been increasingly used in the energy-aware real-time applications on the embedded systems platforms \cite{akusok2019spiking, skatchkovsky2020federated, guan2020unsupervised}. Due to its event-driven nature, \ac{SNN} has the potential to be more energy-efficient than the \ac{ANN}. However, the \ac{SNN} architectures that achieve the high accuracy are usually deep and large, consisting of multiple layers and thousands of neurons and connections \cite{diehl2015fast, kheradpisheh2018stdp, rathi2020enabling, guan2020unsupervised}. For example, the \ac{SNN} architecture proposed in \cite{diehl2015fast} consists of 6 layers with more than 14,500 neurons and 29,000 connections, and achieved an accuracy of 99.14\% on the MNIST dataset. The authors of \cite{rathi2020enabling} implemented an \ac{SNN} based on the VGG-16 architecture \cite{simonyan15}, which consists of more than 138 million connections, and achieved an accuracy of 65.19\% on the ImageNet dataset. The large-scale \acp{SNN} require long computation time and large amount of hardware resources, which limits their potential to be realized on the embedded systems platforms \cite{sen2017approximate, wang2018fpga}. Therefore, we are motivated to explore the techniques to compress the \ac{SNN} to optimize the computation time, hardware resources, and energy consumption on the neuromorphic hardware. Our work aims to improve both the learning and the inference time of the \ac{SNN} while minimizing the loss in the accuracy.
\\\hspace*{1em}There are two common approaches to convert the continuous inputs to the spike events: rate-based coding and temporal coding. In the rate-based coding, the input value is encoded into the spike frequency. An input that has a large value is converted to a series of spike events that occur frequently. On the other hand, in the temporal coding, the input value is encoded into the spike timing, and each neuron spikes at most once for every input. An input that has a large value is converted to a spike event that occurs early. As compared to the rate-based coding, the temporal coding consumes less energy as a fewer number of spikes are generated \cite{rullen2001rate}. In the temporal coding, the input information can be encoded into the relative order (rank order coding) or the latency (\ac{TTFS} coding) of the spikes. The \ac{TTFS} coding has been used in many recent works, such as the ones in \cite{kheradpisheh2018stdp, kheradpisheh2020, mozafari2018first}, to solve challenging real-world problems. Therefore, our work focuses on developing a connection pruning algorithm for the \ac{SNN} with the \ac{TTFS} coding, which is more energy-efficient than the rate-based coding.
\\\hspace*{1em}Connection pruning is a compression technique that has been applied to \ac{ANN} \cite{lecun1990optimal, hassibi1993second} and \ac{SNN} \cite{iglesias2005dynamics, sen2017approximate, rathi2018stdp, chen2018fast, shi2018neuroinspired, shi2019soft, tang2019spike, deng2019comprehensive, kundu2021spike} to reduce the network complexity and energy consumption. It has been shown that more than half of the connections in a well-performing neural network can be removed with minimal impact on the classification accuracy \cite{lecun1990optimal, iglesias2005dynamics}. The connection pruning can be performed either on a pre-trained network \cite{sen2017approximate, chen2018fast} or during the network learning \cite{rathi2018stdp, shi2018neuroinspired, shi2019soft, tang2019spike}. The authors of \cite{sen2017approximate} have proposed a heuristic connection pruning algorithm for the pre-trained \ac{SNN}, in which the connection weights are obtained by converting from those of an \ac{ANN}. The connection pruning is triggered periodically during the inference stage, based on the neuron parameters such as spike rate, membrane potential, and connectivity. Although this approach has efficiently improved the number of operations and hardware energy during the inference stage, it cannot be applied to the applications that require the on-chip learning capability on hardware \cite{sahoo2018online, akusok2019spiking, skatchkovsky2020federated}. In addition, the work in \cite{sen2017approximate} focuses on the \ac{SNN} with the rate-based coding, in which a neuron spikes multiple times during the feed-forward computation of an input image. The spike rate indicates the activity of the neurons and varies greatly among the neurons in the network. In this case, it can be used as a pruning parameter to remove a higher percentage of connections from the less active neurons. However, in the \ac{TTFS} coding, each neuron spikes at most once in the feed-forward computation. Consequently, the spike rate does not vary as much as in the rate-based coding and cannot be used as a pruning parameter. For this reason, the works in \cite{sen2017approximate} and \cite{shi2018neuroinspired, shi2019soft, tang2019spike}, which depend on the spike rate of the neurons to prune the connections, are not applicable to the \ac{SNN} with the energy-efficient \ac{TTFS} coding. Therefore, we proposed a connection pruning approach that is applicable to the energy-efficient \ac{TTFS} coding. In addition, our approach can be applied during the \ac{STDP}-based learning to support the applications that require the on-chip learning capability.
\\\hspace*{1em}Previously, the authors of \cite{rathi2018stdp} have proposed an algorithm that skips the membrane potential update when the connection weight is below a threshold during the \ac{STDP}-based learning. The connection is not eliminated from the network; it can still participate in the \ac{STDP}-based learning and has a chance to be strengthened in the future. Completely eliminating the connections may lead to pre-mature pruning (as these connections may be strengthened later during the \ac{STDP}-based learning) and cause accuracy degradation. However, it helps to reduce not only the number of membrane potential updates but also the number of weight updates during the \ac{STDP}-based learning. Therefore, we are motivated to develop a connection pruning approach that is capable of eliminating the connections completely during the \ac{STDP}-based learning while minimizing the accuracy loss.
\\\hspace*{1em}In this work, we propose a novel connection pruning approach to prune the \ac{SNN} with the \ac{TTFS} coding during the on-chip \ac{STDP}-based learning. As will be discussed in Section \ref{sec:results}, our approach can improve the learning time by 2.1x, reduce the network connectivity by 92.83\%, and save 64\% energy consumption during the on-chip learning without causing accuracy degradation. As compared to our baseline (without pruning), we achieved 2.83x speed-up and 78.24\% reduction in inference time and energy consumption, respectively. Furthermore, our hardware implementation for the connection pruning incurs as little as 0.56\% overhead in the power consumption as compared to our baseline hardware implementation with the pruning units removed. Crucially, our proposed connection pruning approach can be applied during the on-chip \ac{STDP}-based learning for deep \ac{SNN} with the \ac{TTFS} coding. Note that it is more challenging to apply the connection pruning on a deep \ac{SNN} than on a one-layer \ac{SNN} as the learning errors may propagate through the network layers, causing more spuriousness in the deep layers. It is also more challenging to compress the \ac{SNN} with the \ac{TTFS} coding as compared to the rate-based coding. In the \ac{TTFS} coding, there are few spikes emitted, as one neuron emits at most one spike. Removing the connections, which causes the number of spikes to decrease, may lead to insufficiency of spikes to activate the spike activities in the deep layers. 
\section{Preliminaries} \label{sec:model}
\begin{figure}[ht]
	\centering
	\includegraphics[width=.8\linewidth]{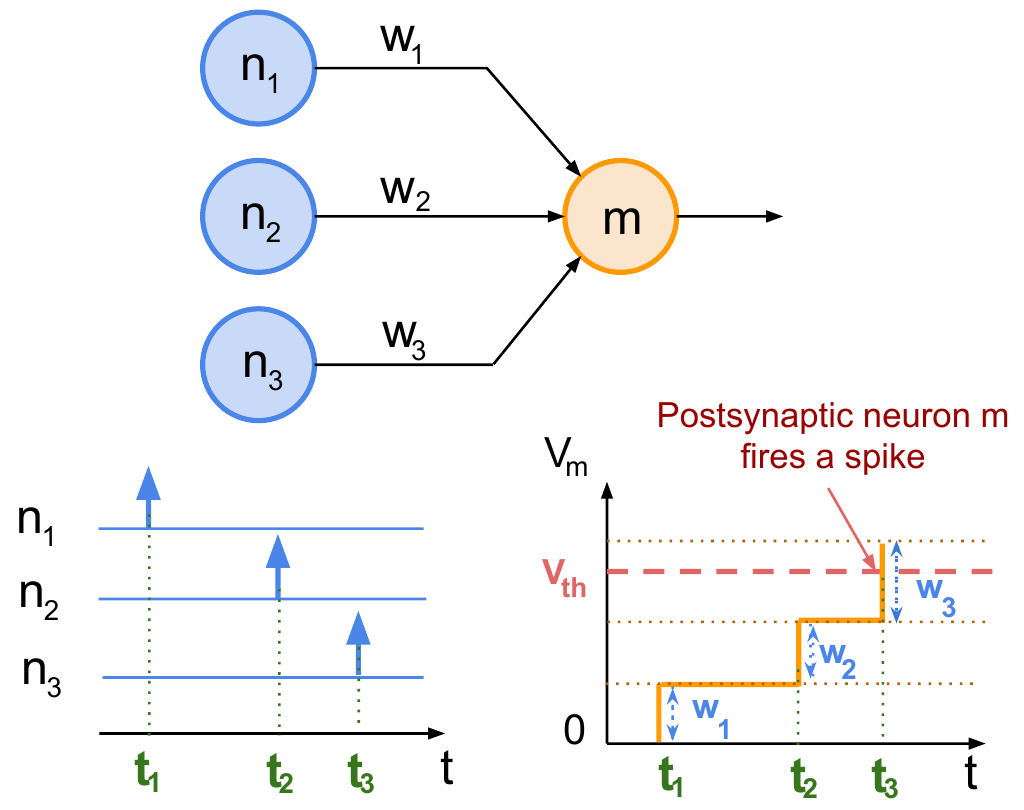}
	\caption{An example of the potential accumulation in the postsynaptic neuron ${m}$ connected to three presynaptic neurons $n_{1}$, $n_{2}$, and $n_{3}$. The connection weights are $w_{1}$, $w_{2}$, and $w_{3}$ respectively. When there is a spike from one of the presynaptic neurons, the membrane potential $V_{m}$ of neuron $m$ is increased by the corresponding weight. When $V_{m}$ exceeds the firing threshold $V_{th}$, neuron $m$ emits a postsynaptic spike.}
	\Description{An example of the spike activities of three presynaptic neurons $n_1$, $n_2$, $n_3$ and the membrane potential update in the postsynaptic neuron $m$. When the spikes arrive at the presynaptic neuron, the membrane potential $V_m$ of the postsynaptic neuron is increased by the corresponding connection weight. When $V_m$ exceedes a pre-defined firing threshold, the postsynaptic neuron $m$ emits a spike.}
	\label{fig:snn}
\end{figure}
In \ac{SNN}, the inputs are discrete spike events. When a spike event arrives at a neuron, the corresponding synaptic weight is accumulated in the membrane potential of the neuron. In our work, the potential accumulation follows the \ac{IF} model \cite{ermentrout1996type}, formulated in the following equation:
\begin{equation}
v_{i}(t) = v_{i}(t-1) + \sum_{j}w_{j,i}s_{j}(t-1)
\end{equation}
where
\begin{equation}
s_{i}(t) = 
\begin{cases}
1 &\mbox{if neuron $i$ spikes at time $t$}\\
0 & \mbox{otherwise}
\end{cases}
\end{equation}
$v_{i}(t)$ and $s_{i}(t)$ are the membrane potential and postsynaptic spike of neuron $i$ at time $t$, $w_{j,i}$ is the synaptic weight between neuron $i$ and neuron $j$. The overview of this membrane potential accumulation is illustrated in Figure \ref{fig:snn}. As soon as the membrane potential exceeds the firing threshold, the neuron fires a postsynaptic spike and inhibits the other neurons that are close to it in the network.
\\\hspace*{1em}The \ac{STDP}-based learning algorithm in \ac{SNN} is performed based on the correlation between the presynaptic spike and postsynaptic spike of a connection. If the postsynaptic spike occurs within a small time window from the presynaptic spike, the two spikes are considered correlated and the connection is strengthen. This process is called \ac{LTP}. Otherwise, the two spikes are considered uncorrelated and the connection is weakened. This process is called \ac{LTD} \cite{song2000competitive}. In our work, we applied a hardware-friendly \ac{STDP}-based learning algorithm, as proposed in the work in \cite{kheradpisheh2018stdp}. In this algorithm, the synaptic weights are updated based on the following Equation:
\begin{equation}
\Delta w = 
\begin{cases}
a^{+}w_{ij}(1-w_{ij}) &\mbox{if } t_{j}-t_{i} \leq 0\\
a^{-}w_{ij}(1-w_{ij}) &\mbox{otherwise}
\end{cases}
\end{equation}
where $a^{+}$ and $a^{-}$ are the learning rates, $t_{j}$ and $t_{i}$ are the time of the presynaptic and postsynaptic spikes, respectively. The values of the connection weights are in the range [0, 1]. The \ac{STDP}-based learning is performed layer by layer: a layer starts its learning only after the preceding layer finishes its learning. In the next section, we will describe our connection pruning approach applied to this \ac{SNN} model and \ac{STDP}-based learning algorithm. Note that our work focuses on the network compression technique rather than the neuron model and the learning algorithm, which were adopted from the work in \cite{kheradpisheh2018stdp}.
\section{Proposed Connection Pruning Approach} \label{sec:method}
\begin{figure*}[t]
	\centering
	\includegraphics[width=.9\linewidth]{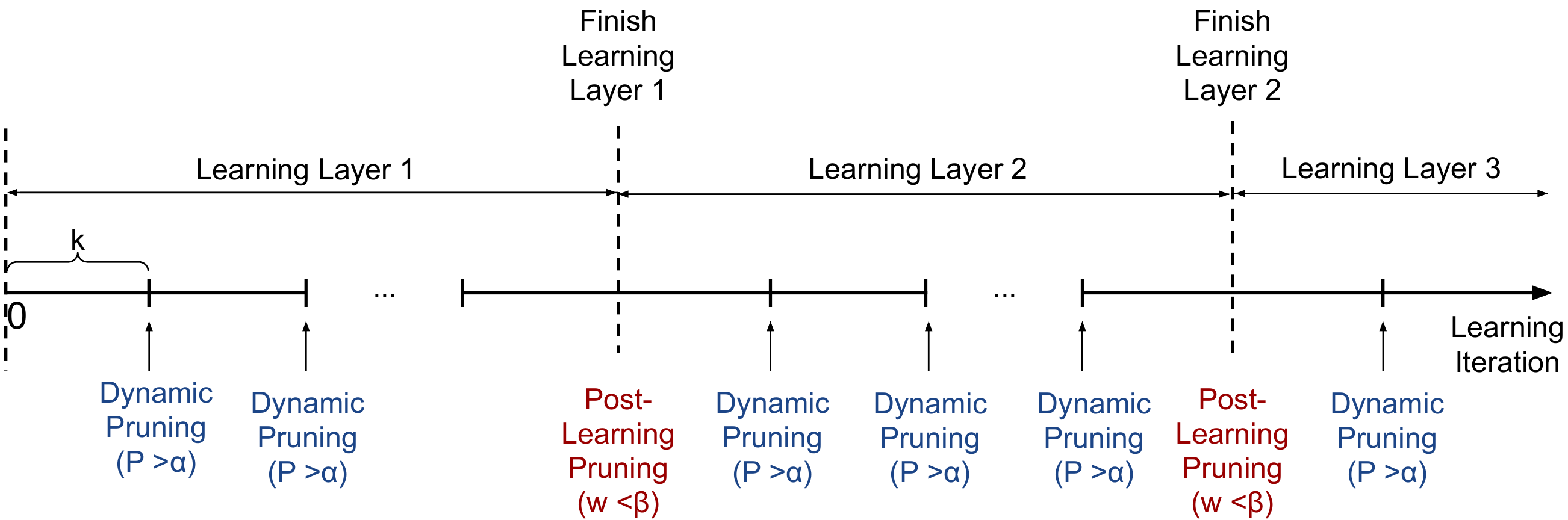}
	\caption{Overview of our two-stage connection pruning approach.}
	\Description{A diagram that explains our two-stage connection pruning approach. During the \ac{STDP}-based learning, every connection is associated with a prune parameter $P$, as described in Equation \ref{eq:p}. After every $k$ iteration, the connections that have $P$ greater than a threshold $\alpha$ are eliminated (i.e. set to zero value). This process is refered to as dynamic pruning. In addition, when the learning in each layer has finished, the connections that have the weights less than a threshold $\beta$ are eliminated. This process is refered to as post-learning pruning.}
	\label{fig:method_overview}
\end{figure*}
In order to compress the \ac{SNN} and reduce the on-chip \ac{STDP}-based learning time, we propose a connection pruning approach consisting of two stages: (i) dynamic pruning during the on-chip \ac{STDP}-based learning in every layer and (ii) post-learning pruning after each layer has learned, as shown in Figure \ref{fig:method_overview}. Note that both of these stages are performed during the on-chip \ac{STDP}-based learning.
\subsection{Dynamic Pruning During The On-chip STDP-Based Learning}
In the dynamic pruning stage, the connection pruning is performed after every $k$ iterations during the on-chip \ac{STDP}-based learning in each layer, based on two parameters: (i) weight update history, $h$, of the synaptic connection and (ii) time of postsynaptic spike, $t_{post}$. In the following, we will explain these parameters and present the details of our dynamic pruning approach.
\subsubsection{Weight Update History}
\begin{figure}[t]
	\centering
	\begin{subfigure}{.49\linewidth}
		\includegraphics[width=\linewidth]{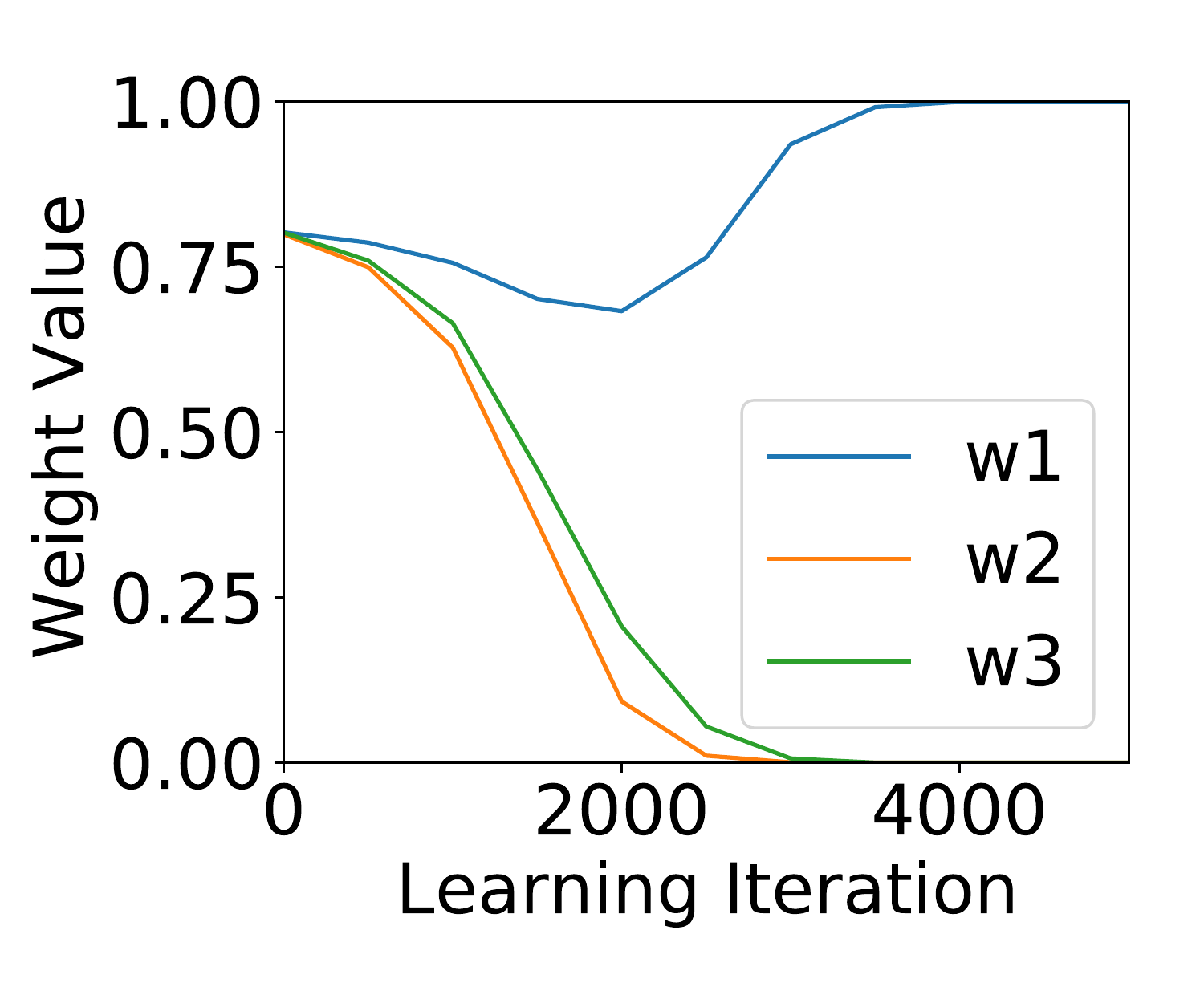}
		\caption{}
		\Description{Weight values in the range [0, 1] throughout the learning iterations, obtained from a run on our hardware implementation with the connection pruning turned off. There are three connection weights $w_1$, $w_2$, and $w_3$. $w_1$ and $w_2$ converge to 0 while $w_3$ converge to 1 after the learning is finished.}
		\label{fig:method_weights_noprune}
	\end{subfigure}
	\begin{subfigure}{.49\linewidth}
		\includegraphics[width=\linewidth]{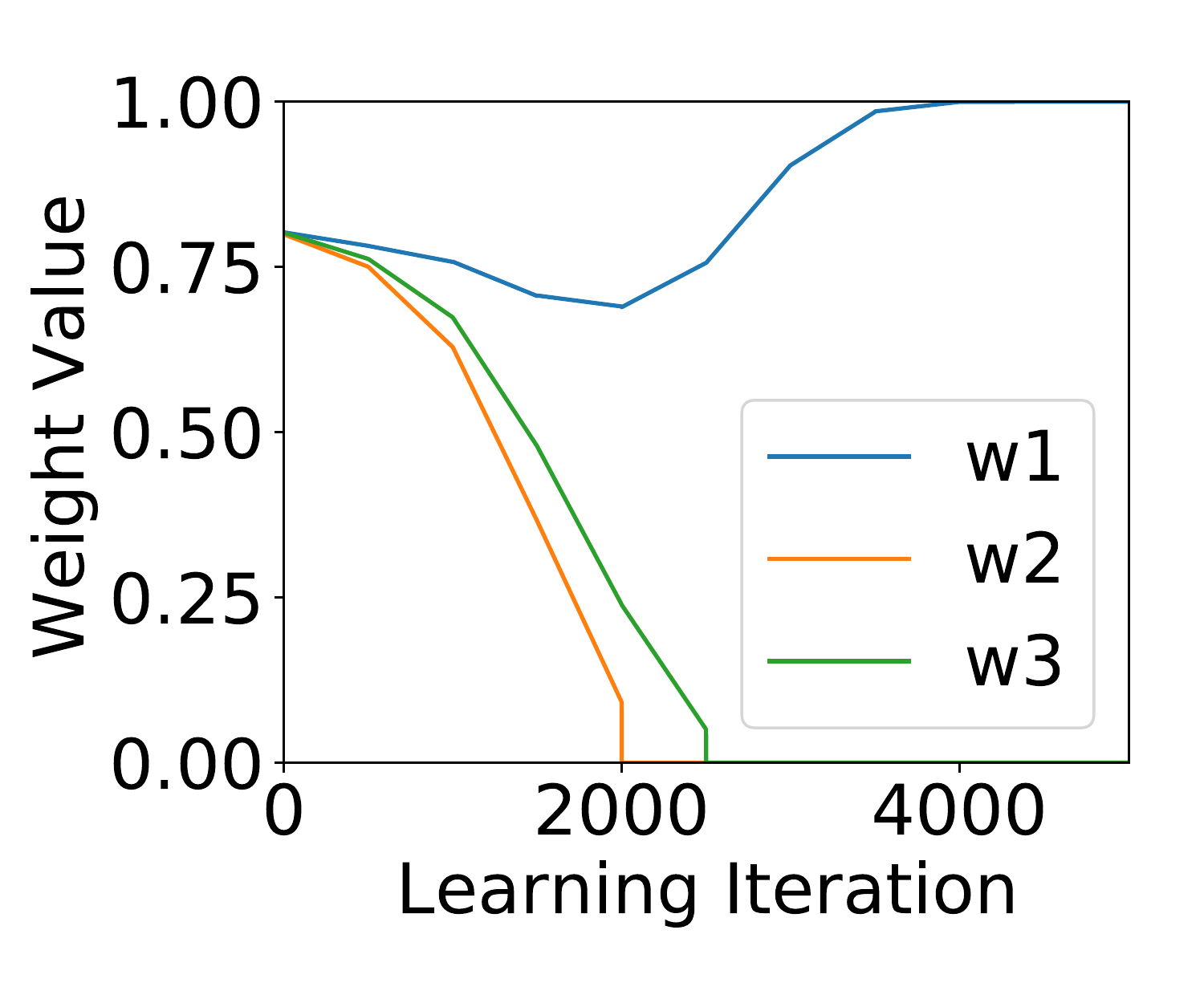}
		\caption{}
		\Description{Weight values in the range [0, 1] throughout the learning iterations, obtained from a run on our hardware implementation with the connection pruning turned on. The connection weights $w_1$ and $w_2$ decrease steeply and reach a low value at iteration 2000 and 2500, respectively. These connections are eliminated (i.e. set to zero value) at iteration 2000 and 2500.}
		\label{fig:method_weights_pruned}
	\end{subfigure}
	\caption{Weight values of selected connections throughout the iterations of the on-chip \ac{STDP}-based learning performed on our hardware implementation (a) without connection pruning (b) with connection pruning.}
\end{figure}
The weight update history, $h$, is defined in the following equation:
\begin{equation}\label{eq:h}
h = \frac{d}{w}
\end{equation}
where $d$ is the number of times the \ac{LTD} is performed on the connection, or the number of times the connection weight is decreased, in $k$ learning iterations. $w$ is the weight value at the time of pruning. The goal is to prune the connections having weights that are (i) small and (ii) decreasing steeply throughout a number of learning iterations. Note that the connection weights of the \ac{SNN} emulated in our work are in the range [0, 1]. The intuition is that if a connection weight, which is in the range [0, 1], has been decreasing steeply in a time period and has reached a small value, it will likely continue to decrease until it approaches zero. Therefore, we propose to prune these connections early to reduce the number of membrane potential updates and learning computations in the future iterations. For example, in Figure \ref{fig:method_weights_noprune}, which shows the learning progress of selected weights during the on-chip \ac{STDP}-based learning performed on our hardware implementation, weights $w_2$ and $w_3$ decrease steeply to a small value (less than 0.1) from iteration 2000 to 2500 and 1500 to 2000, respectively. In the later iterations, these weights contribute little to the spiking activities and will approach zero eventually. Consequently, $w_2$ can be pruned at iteration 2000 and $w_3$ can be pruned at iteration 2500, as shown in Figure \ref{fig:method_weights_pruned}. Note that having the weight value in the denominator helps to reduce the chance of pre-mature pruning, in which the connection is pruned when it still contributes non-negligibly to the spiking activities.
\subsubsection{Time of Postsynaptic Spike}
\begin{figure}[t]
	\centering
	\begin{subfigure}{.49\linewidth}
		\centering
		\includegraphics[width=.9\linewidth]{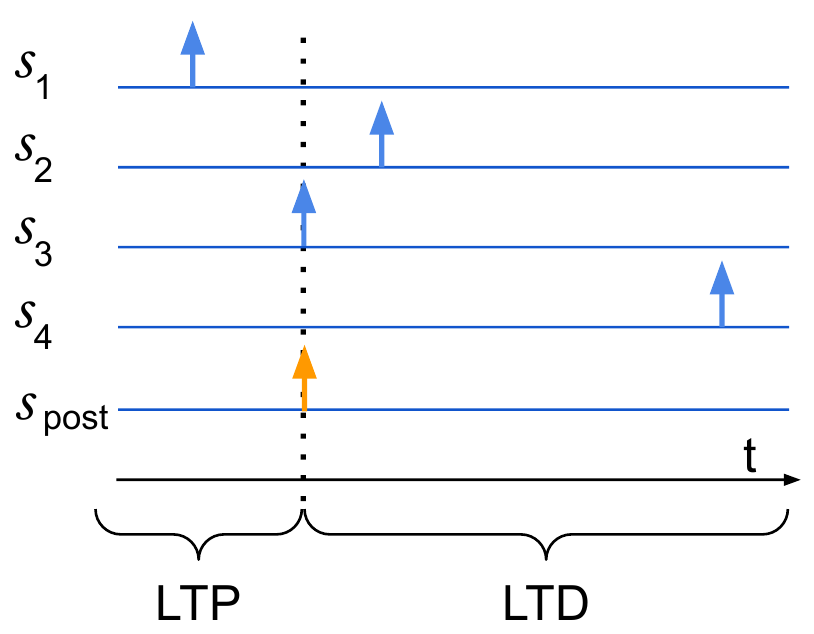}
		\caption{}
		\Description{An example of the spikes at four presynaptic neurons $s_{1}, s_{2}, s_{3}, s_{4}$ and a postsynaptic neuron $s_{post}$. $s_1$ spikes before $s_{post}$, $s_3$ spikes at the same time as $s_{post}$, and $s_{2}$ and $s_{4}$ spike after $s_{post}$.}
		\label{fig:method_spike_early}
	\end{subfigure}
	\begin{subfigure}{.49\linewidth}
		\centering
		\includegraphics[width=.9\linewidth]{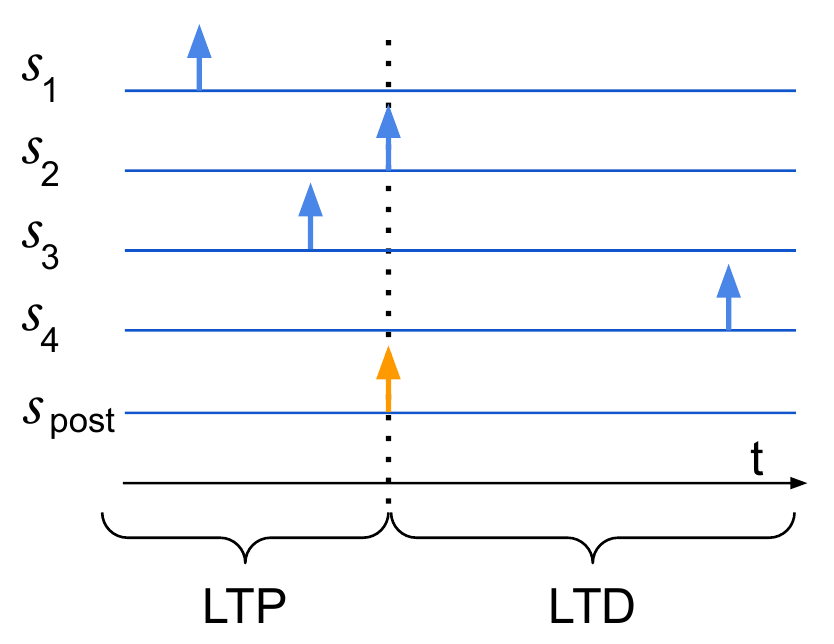}
		\caption{}
		\Description{An example that shows similar spike activities as Figure \ref{fig:method_spike_early} except $s_{post}$ spikes at the same time as $s_2$.}
		\label{fig:method_spike_late}
	\end{subfigure}
	\caption{Examples of the spiking activities of four presynaptic neurons $s_{1}, s_{2}, s_{3}, s_{4}$ and a postsynaptic neuron $s_{post}$. (a) $s_{post}$ spikes earlier, more connection weights are decreased (b) $s_{post}$ spikes later, fewer connection weights are decreased.}
	\label{fig:method_spike_activity}
\end{figure}
In addition to the weight update history, our proposed pruning approach considers the synaptic competition during the on-chip \ac{STDP}-based learning to formulate the pruning decision. The connection weights that are decreased during a strong synaptic competition should be less likely to be pruned than those that are decreased during a synaptic competition that is not as strong. The strength of the synaptic competition can be estimated using the timing of the postsynaptic spikes. We will explain this relationship in the following.
\\\hspace*{1em}In our implementation of the on-chip \ac{STDP}-based learning, the connection weights are increased if the presynaptic neurons spike before the postsynaptic neurons; otherwise, the connection weights are decreased, as described in Section \ref{sec:model}. If there are many large connection weights, few spike events are needed to cause the membrane potential of the postsynaptic neuron to exceed the firing threshold. Hence, only the most competitive connections that have the largest weight values and carry the earliest presynaptic spikes will be able to contribute to the spiking of the postsynaptic neuron. Figure \ref{fig:method_spike_early} shows an example to illustrate this scenario. When the postsynaptic spike, $s_{post}$, is emitted early, it is only the presynaptic spikes $s_{1}$ and $s_{3}$ that were early enough to contribute to $s_{post}$ and their connection weights will be increased. Contrarily, although $s_{2}$ is relatively early as compared to $s_{4}$ and its connection weight may be large, it was not able to contribute to $s_{post}$. Consequently, the connection weight associated with $s_{2}$ will be decreased. On the other hand, if $s_{post}$ is emitted later, as shown in Figure \ref{fig:method_spike_late}, $s_{2}$ will be able to contribute to $s_{post}$ and its connection weight will be increased. Therefore, the weight decrements in the former scenario (Figure \ref{fig:method_spike_early}) should be weighted less than the weight decrements in the latter scenario (Figure \ref{fig:method_spike_late}) in the connection pruning decision.
\\\hspace*{1em}The strong synaptic competition, as explained above, may happen in the early iterations in the on-chip \ac{STDP}-based learning on our hardware, as shown in Figure \ref{fig:method_weights_noprune}. We initialized the connection weights to follow the normal distribution in the range of (0, 1) with the mean being skewed at 0.8, following the work in \cite{sdnnpython}. This is to encourage the spiking activities and to emphasize the most dominant presynaptic spikes and connections in the early iterations of the learning. The connections that have smaller weights or carry less dominant presynaptic spikes will have their weights decreased. However, these connections still have a chance to have their weights increased in a later learning iteration, when many of the other connections are weakened and the postsynaptic neurons do not spike as early as before, as shown in weight $w_1$ in Figure \ref{fig:method_weights_noprune}. Therefore, the time of postsynaptic spikes, $t_{post}$, helps to avoid pruning these connections in the early learning iterations when the synaptic competition is strong.
\subsubsection{The Dynamic Pruning Approach} 
The proposed dynamic pruning approach combines $h$ and $t_{post}$ into a pruning parameter $P$:
\begin{equation}
P = h * t_{post} \label{eq:hc}
\end{equation}
Combining equations \eqref{eq:h} and \eqref{eq:hc}, we have:
\begin{equation}
P = \frac{d}{w} * t_{post} \label{eq:p}
\end{equation}
$P$ is evaluated for every connection in every $k$ iterations of the on-chip \ac{STDP}-based learning. If $P$ is greater than a pre-defined threshold $\alpha$, the connection is pruned. This dynamic pruning stage is performed during the on-chip \ac{STDP}-based learning in each layer. After the learning is finished, we proceed to the post-learning pruning stage, which we will describe next.
\subsection{Post-Learning Pruning}
Our post-learning pruning approach is performed after the learning in one layer has finished, before proceeding to the next layer, as shown in Figure \ref{fig:method_overview}. In the post-learning pruning stage, all the connections that have the weights less than a threshold, $\beta$, are eliminated. Note that although this stage is performed after the learning in a layer, it is still within the on-chip learning process of the \ac{SNN}. It differs from the existing pruning approaches on the pre-trained networks as it can affect the learning in the next layer. Overly aggressive pruning of the preceding layer will lead to learning errors in the next, causing losses in the classification accuracy.
\section{Evaluation Hardware Architecture} \label{sec:architecture}
\begin{figure}[t]
	\centering
	\includegraphics[width=\linewidth]{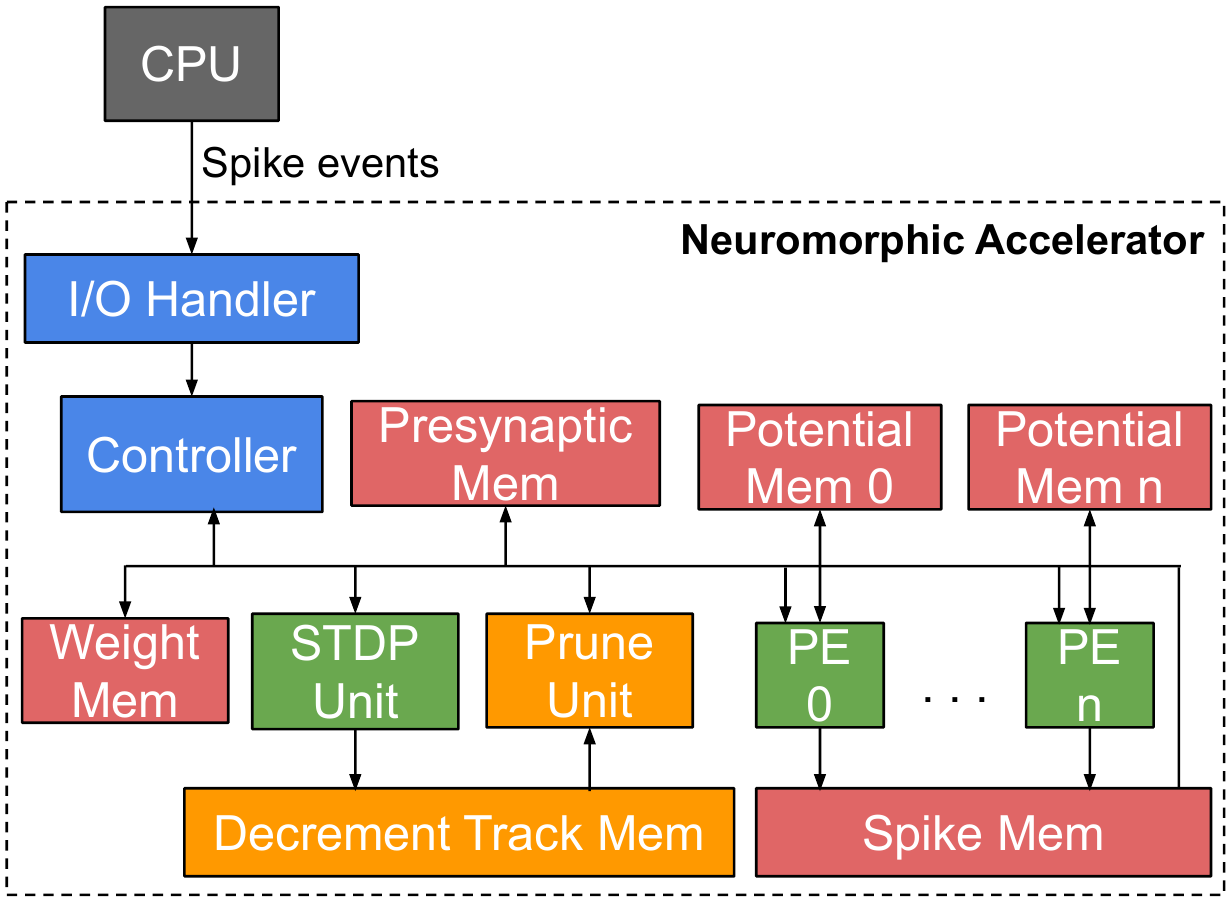}
	\caption{Our evaluation hardware architecture, which incorporates our connection pruning approach to reduce the on-chip learning time and the network connectivity for \ac{SNN}.}
	\Description{Our evaluation hardware architecture. The spike events are received by the Controller, which is connected to the Weight Memory, the STDP Unit, the Prune Unit, the Decrement Track Memory, the Spike Memory, and an array of PEs. Each PE is connected to a Potential Memory. The STDP Unit and the Prune Unit are connected to the Decrement Track Memory.}
	\label{fig:architecture}
\end{figure}
We designed an event-driven hardware architecture that incorporates our proposed connection pruning approach to reduce the on-chip learning time and network connectivity for \ac{SNN}, as shown in Figure \ref{fig:architecture}. The overall flow of our hardware architecture is as follows. In every time step, the Controller receives the spike events and forwards them to the \acp{PE}. The \acp{PE} compute the membrane potentials and update the Potential Memory. At the end of the time step, the \acp{PE} compares each membrane potential in the Potential Memory with a pre-defined firing threshold. If the membrane potential of a neuron exceeds the firing threshold, the \ac{PE} writes the corresponding neuron ID to the Spike Memory. Finally, the Controller activates the STDP unit to perform the weight updates based on the \ac{STDP}-based learning algorithm that was described in Section \ref{sec:model}. During the learning, the STDP unit records the decrements of the weight values of each connection in the Decrement Track Memory.
\\\hspace*{1em}The connection pruning is performed in the Prune Unit in every $k$ iterations. After the \ac{STDP} unit finishes updating the connection weights for the current time step, the Prune unit evaluates the pruning criteria for each connection weight (described in Section \ref{sec:method}). If the pruning criteria is met, the corresponding connection weight is set to zero in the Weight Memory. The pruning components incur little overheads in the resources and power consumption in our hardware implementation which we will discuss in the following sections.
\section{Experimental Setup} \label{sec:expsetup}
This section describes our experimental setup to evaluate our proposed connection pruning approach and hardware architecture. We will first present the hardware platform on which our hardware architecture was implemented, followed by the dataset and the \ac{SNN} network parameters used in our experiments. 
\subsection{Hardware Platform}
Our hardware architecture was implemented on the Zynq-7000 Zedboard (XC7Z020) using Verilog. The hardware resources and power consumption were estimated by Xilinx Vivado 2018.3 \cite{feist2012vivado}. Our hardware implementation of the connection pruning incurs less than 3.52\% overheads in the LUTs and FFs and approximately 8.7\% overheads in the BRAM consumption as compared to our baseline hardware implementation (with the Prune Unit and the Decrement Track Memory removed). In addition, 11 DSP slices were used in the implementation of the connection pruning, which is 5\% of the DSP slices available on the board. The power consumption was increased by 0.56\% as compared to our baseline hardware implementation (no pruning). These overheads are reported based on the post-implementation estimation generated by Xilinx Vivado 2018.3. Our hardware implementation was run at the clock frequency 100 MHz.
\subsection{Dataset and Network Configuration} \label{sec:dataset}
Our connection pruning approach was evaluated on the Caltech 101 dataset \cite{fei2004learning}. The training set consists of 400 images of the two categories: (i) human face and (ii) motorbike in 160x250 pixels. The test set consists of 396 images. Our hardware implementation of the \ac{SNN} was based on \cite{kheradpisheh2018stdp} and \cite{sdnnpython}. The \ac{SNN} implemented on our hardware consists of three convolutional-pooling layers. The kernels of the three convolutional layers are of size 5x5x4, 17x17x20, and 5x5x20 (width x height x depth). In between the convolutional layers, the downsampling was performed with window sizes 7x7 and 5x5, respectively. Our connection pruning approach was applied to the convolutional layers. Following to the work in \cite{kheradpisheh2018stdp}, the images were pre-processed using a \ac{DoG} filter, followed by the \ac{SNN} emulated on our hardware implementation, and classified using a \ac{SVM}. Note that our work does not focus on improving this model but developing a connection pruning algorithm to reduce the network complexity of the \ac{SNN} architecture. The effects of our connection pruning approach will be discussed in the following section.
\section{Results and Discussion} \label{sec:results}
The performance of our connection pruning approach was evaluated based on the following four metrics: (i) the number of connections reduced in the network, (ii) the time speed-up, (iii) the energy saved, and (iv) the classification accuracy. In this section, we will discuss the trade-offs between these metrics.  In addition, the individual effects achieved by each of the stages: (i) dynamic pruning and (ii) post-learning pruning in our two-staged connection pruning approach will be analysed. Finally, we will discuss the impacts of connection pruning on the network behaviours in the inference stage. We will demonstrate that our connection pruning approach successfully eliminates a significant number of connections during the on-chip \ac{STDP}-based learning, which helps to reduce the time and energy in both the learning and the inference stages, without incurring any accuracy loss.
\subsection{Connectivity Reduction vs. Accuracy}
\begin{figure}[t]
	\centering
	\includegraphics[width=.9\linewidth]{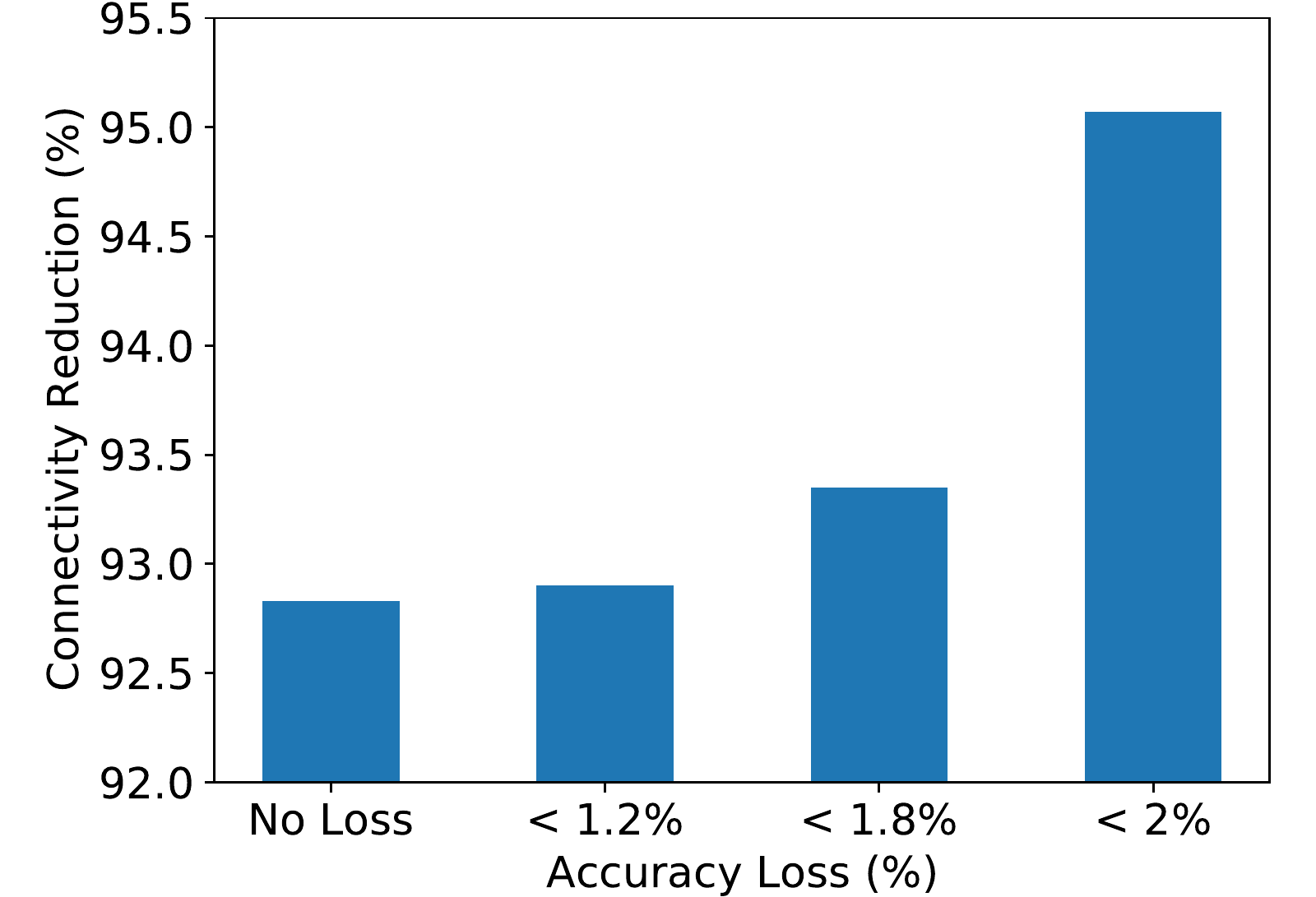}
	\caption{Connectivity reduction and accuracy trade-offs. The connectivity reduction was computed as $n_{pruned}$/$n_{total}$, where $n_{pruned}$ is the number of connections being pruned and $n_{total}$ is the total number of connections in the network.}
	\Description{Connectivity reduction for different trade-offs in accuracy. The connectivity is reduced by 92.83\%, 92.9\%, 93.35\%, and 96.2\% when the accuracy loss is allowed to be up to 0\%, 1,2\%, 1.8\%, and 2\%, respectively.}
	\label{fig:conn_acc}
\end{figure}
\begin{table*}[t]
	\centering
	\caption{Classification accuracy achieved on the Caltech-101 (Face/Motorbike) dataset.}
	\label{tab:comparison_accuracy}
	\begin{tabular}{|c|c|c|c|c|c|c|c|}
		\hline
		& \textbf{\cite{lee2018deep}} & \textbf{\cite{mozafari2018first}} & \textbf{\cite{kheradpisheh2018stdp}} & \textbf{\cite{kheradpisheh2020}} & \textbf{\cite{zhang2020rectified}} & \textbf{Our Work} \\\hline
		SNN Architecture & CNN & CNN & CNN & Fully Connected & CNN & CNN \\\hline
		Learning Algorithm & \ac{STDP} & Reward-modulated \ac{STDP} & \ac{STDP} & Back-propagation & Back-propagation & \ac{STDP} \\\hline
		Platform & GPU & GPU & GPU & GPU & CMOS & FPGA \\\hline
		\# Learnable Parameters & 25,488 & 23,120 & 25,480 & 160,008 & >110 M & \textbf{2,383} \\\hline
		\shortstack{On-Chip Learning\\on Embedded Hardware} & No & No & No & No & No & \textbf{Yes} \\\hline
		Accuracy & 97.6\% & 98.9\% & 99.1\% & 99.2\% & \textbf{99.5\%} & 95.7\% \\\hline
	\end{tabular}
\end{table*}
Our proposed connection pruning approach helps to eliminate 92.83\% connections in the network without incurring any loss in the classification accuracy, as presented in Figure \ref{fig:conn_acc}. This is consistent with the result in \cite{iglesias2005dynamics} that more than 90\% of the connections can be eliminated after one million iterations of the \ac{STDP}-based learning, regardless of the network size. However, in the work in \cite{iglesias2005dynamics}, the connections are eliminated after a large number of iterations (one million iterations). Our connection pruning approach eliminates the connections early to save the on-chip learning time while preserving the accuracy. Different trade-offs between the number of connections eliminated and the accuracy loss are presented in Figure \ref{fig:conn_acc}. The number of connections in the network can be reduced by 92.83\% without incurring any loss in the accuracy. Moreover, when the accuracy is allowed to fall within 1.2\%, the network can be compressed by 92.9\%. When up to 1.8\% loss in the accuracy is allowed, the network connectivity can be reduced by 93.35\%. In addition, when the accuracy loss is maintained within 2\%, the connectivity can be reduced by up to 95.07\%. The proposed connection pruning algorithm, which does not cause any loss in the classification accuracy, helps to reduce the number of connections to 9.7 times-55 million times fewer than other \ac{SNN} implementations in the existing works, as shown in Table \ref{tab:comparison_accuracy}. Note our work does not aim to achieve the highest accuracy but a balance between the accuracy and the network complexity, which has significant impacts on the response time and the energy consumption on embedded systems platforms. In the case our connection pruning algorithm is performed off-chip (no on-chip learning is needed), it can help to reduce the number of entries in the Weight Memory by 92.83\% as compared to our baseline (the \ac{SNN} described in Section \ref{sec:dataset}). 
\subsection{Improvements in Learning Time and Energy Consumption}
\begin{figure}[t]
	\centering
	\includegraphics[width=.9\linewidth]{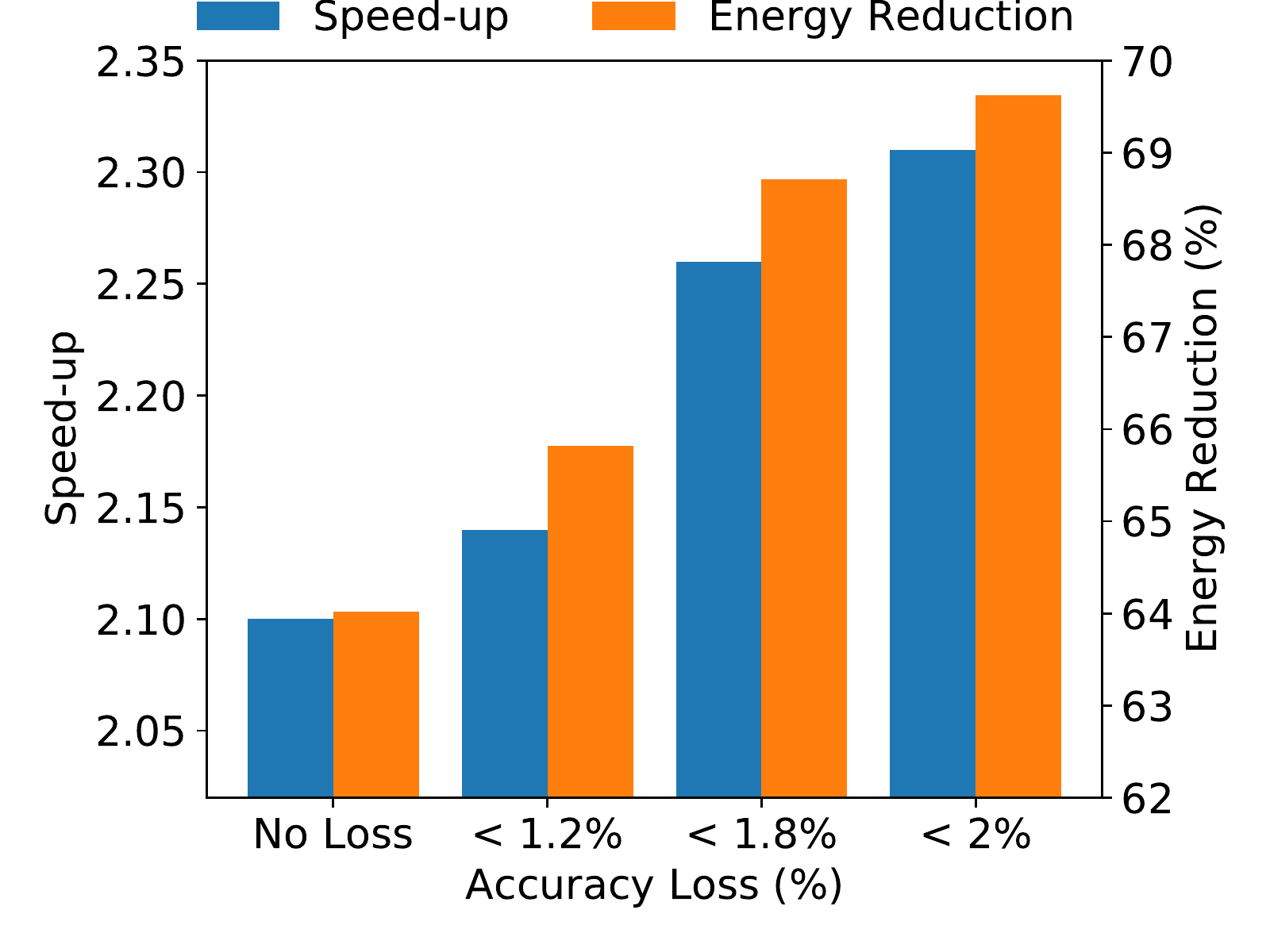}
	\caption{Trade-offs between learning time, energy consumption, and classification accuracy of the \ac{SNN} emulated on our hardware implementation. The learning time speed-up and energy reduction are computed as $t_{noprune}/t_{pruned}$ and $(1-E_{pruned}/E_{noprune})*100$, respectively. $t_{pruned}$ and $t_{noprune}$ are the learning time with and without the connection pruning, $E_{pruned}$ and $E_{noprune}$ are the energy consumption in the learning with and without pruning.}
	\Description{Learning time, energy consumption, and accuracy trade-offs. The learning time is reduced by 2.1x, 2.14x, 2.26x, and 2.31x, and the energy consumption is reduced by 64\%, 65.82\%, 68.71\%, and 69.62 when the accuracy loss is allowed to be up to 0\%, 1.2\%, 1.8\%, and 2\%, respectively.}
	\label{fig:speedup_acc}
\end{figure}
The speed-up in the on-chip \ac{STDP}-based learning time depends on (i) the number of connections being pruned in the network and (ii) the earliness of the pruning during the on-chip learning. Our connection pruning approach helps to speed-up the on-chip learning time by 2.1x without incurring any loss in the classification accuracy, as shown in Figure \ref{fig:speedup_acc}. In addition, when the accuracy loss is up to 1.2\%, the learning time is improved by 2.14x. Furthermore, the learning time can be improved by up to 2.26x and 2.31x when the accuracy is allowed to fall within 1.8\% and 2\%, respectively. In addition to the learning time, the energy consumption during the \ac{STDP}-based learning is also improved as a result of the network compression. In our experiments, the energy saving is estimated based on the number of operations reduced, similar to the related work in \cite{rathi2018stdp}. As shown in Figure \ref{fig:speedup_acc}, the energy consumption is reduced by 64-69.62\% with 0-2\% loss in the accuracy. The improvements on the learning time and the energy consumption were attributed to the two stages in our connection pruning approach: dynamic pruning and post-learning pruning, which we will analyse in the following.
\subsection{Dynamic Pruning vs. Post-Learning Pruning}
\begin{table}[t]
	\centering
	\caption{Impacts of dynamic pruning and post-learning pruning when applied individually and when combined together in our two-staged connection pruning approach.}
	\label{tab:dynpos_compare}
	\resizebox{\linewidth}{!}{%
		\begin{tabular}{|c|c|c|c|c}
			\hline
			& \textbf{\shortstack{Dynamic\\Pruning}} & \textbf{\shortstack{Post-Learning\\Pruning}} & \textbf{\shortstack{Combined\\Approach}}\\\hline
			Connectivity Reduction & 89.95\% & \textbf{92.8\%} & \textbf{92.8\%} \\\hline
			\shortstack{Learning Time\\Speed-up} & \textbf{2x} & 1.78x & \textbf{2.1x} \\\hline
			Energy Reduction & \textbf{62\%} & 48\% & \textbf{64\%} \\\hline
	\end{tabular}}
\end{table}
Each of the two stages: (i) dynamic pruning and (ii) post-learning pruning in our connection pruning approach has its own effects on the network connectivity and the learning time. While the dynamic pruning is performed periodically during the learning in each layer, the post-learning pruning is performed after the learning in the layer has finished, as illustrated in Figure \ref{fig:method_overview}. Therefore, the dynamic pruning can eliminate the connections earlier than the post-learning pruning. When the two approaches are applied separately, the dynamic pruning achieves a higher learning time speed-up and energy reduction, as shown in Table \ref{tab:dynpos_compare}. However, the connectivity reduction achieved by the dynamic pruning is less than the post-learning pruning, as shown in Table \ref{tab:dynpos_compare}. The reason is that during the learning, the network needs to maintain sufficient connections to generate the spike events in order to stimulate the \ac{STDP}-based weight updates. On the other hand, after the learning in the layer has finished, the weights can be pruned without the concerns of the learning activities. Our connection pruning approach combines the dynamic pruning and the post-learning pruning to achieve the high learning time speed-up, high energy saving and significant connectivity reduction.
\subsection{Impact on Inference Stage}
\begin{figure*}[t]
	\centering
	\begin{subfigure}{.49\linewidth}
		\centering
		\includegraphics[width=.9\linewidth]{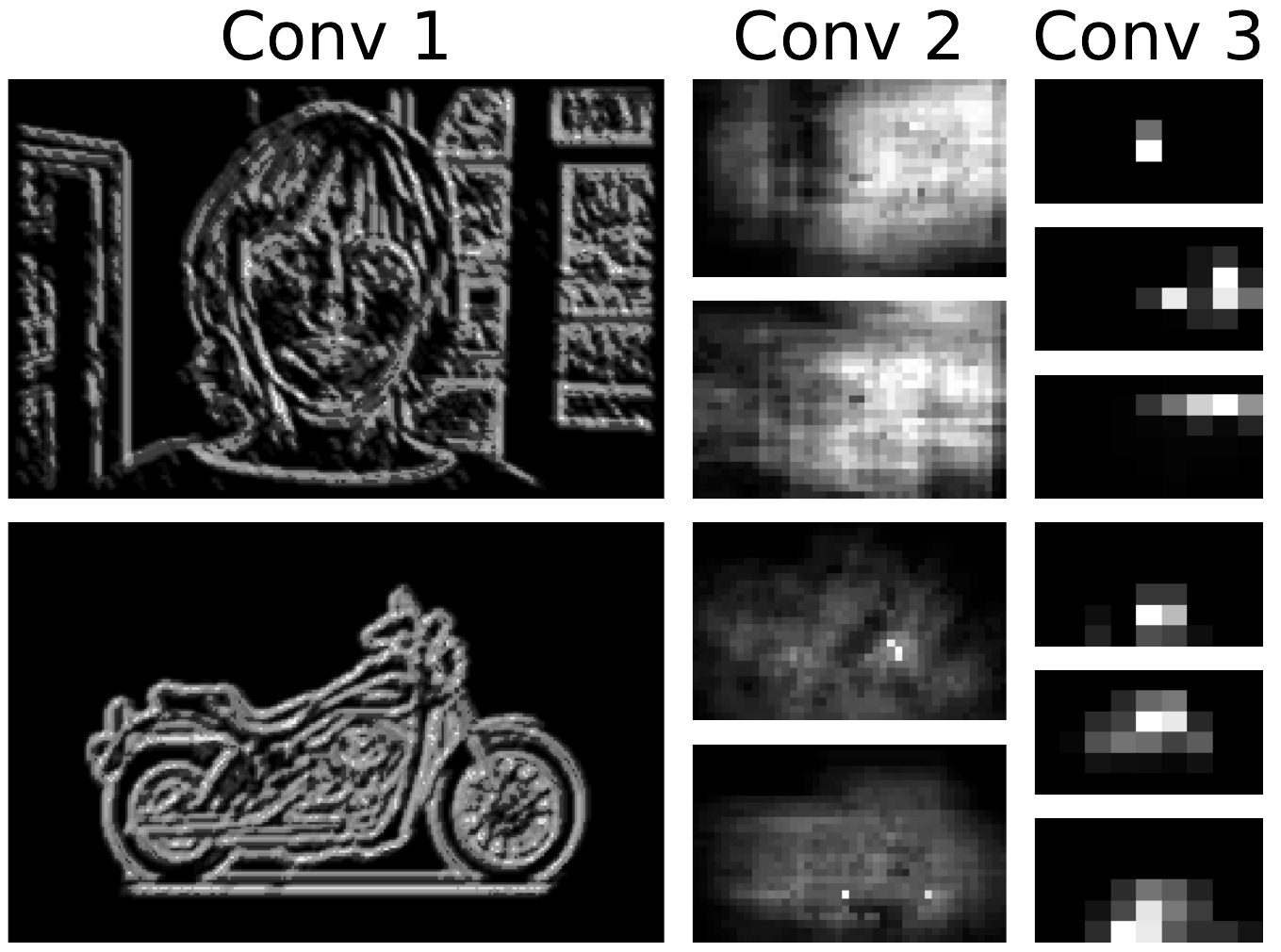}
		\caption{}
		\Description{Selected feature maps in the three convolutional layers with the connection pruning turned off.}
		\label{fig:maps_noprune}
	\end{subfigure}
	\begin{subfigure}{.49\linewidth}
		\centering
		\includegraphics[width=.9\linewidth]{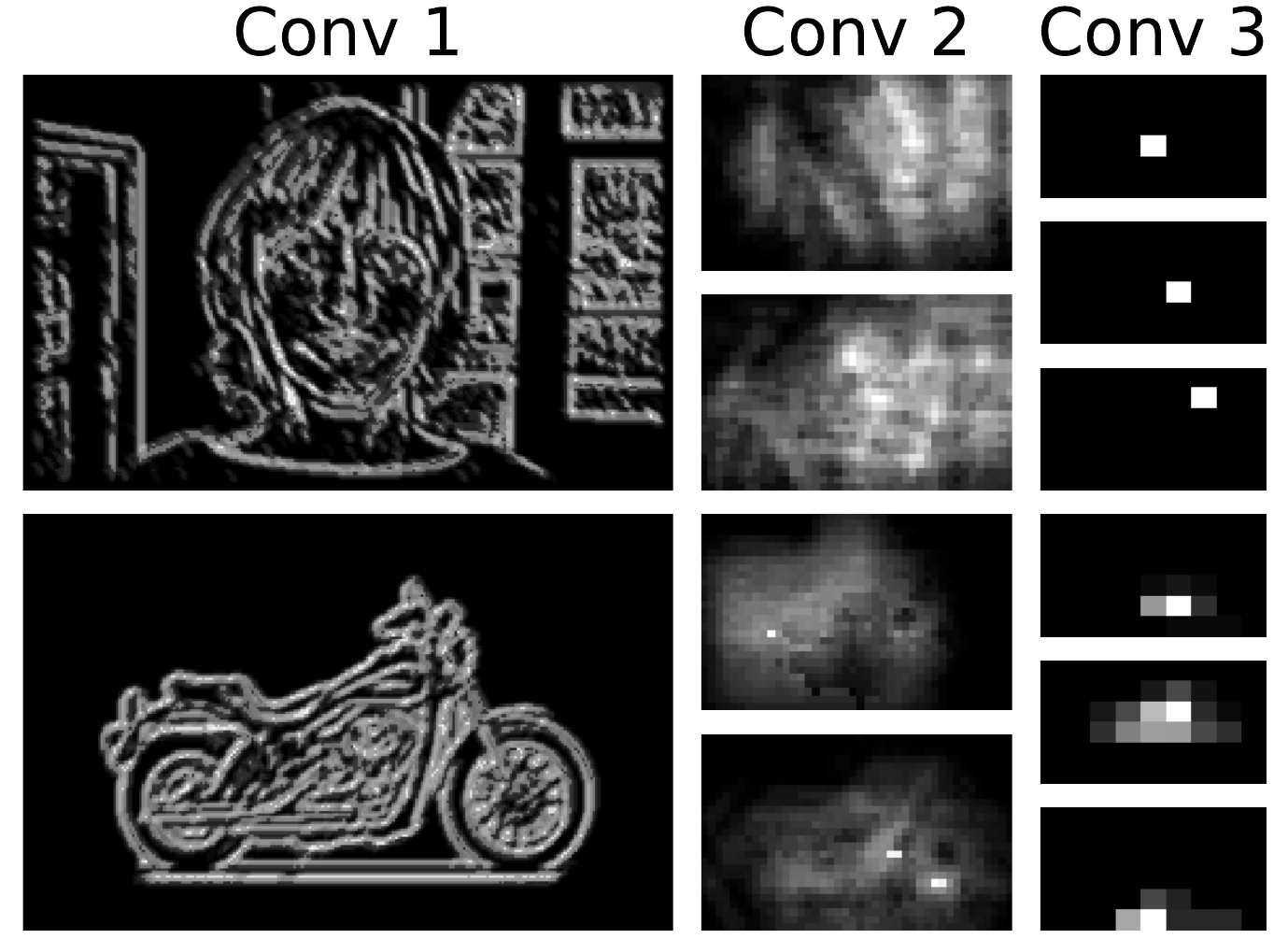}
		\caption{}
		\Description{Selected feature maps in the three convolutional layers with the connection pruning turned on.}
		\label{fig:maps_pruned}
	\end{subfigure}
	\caption{Selected feature maps in different convolutional layers after the \ac{STDP}-learning (a) without connection pruning and (b) with connection pruning on our hardware implementation.}
	\label{fig:featuremaps}
\end{figure*}
\begin{figure}[t]
	\centering
	\includegraphics[width=.9\linewidth]{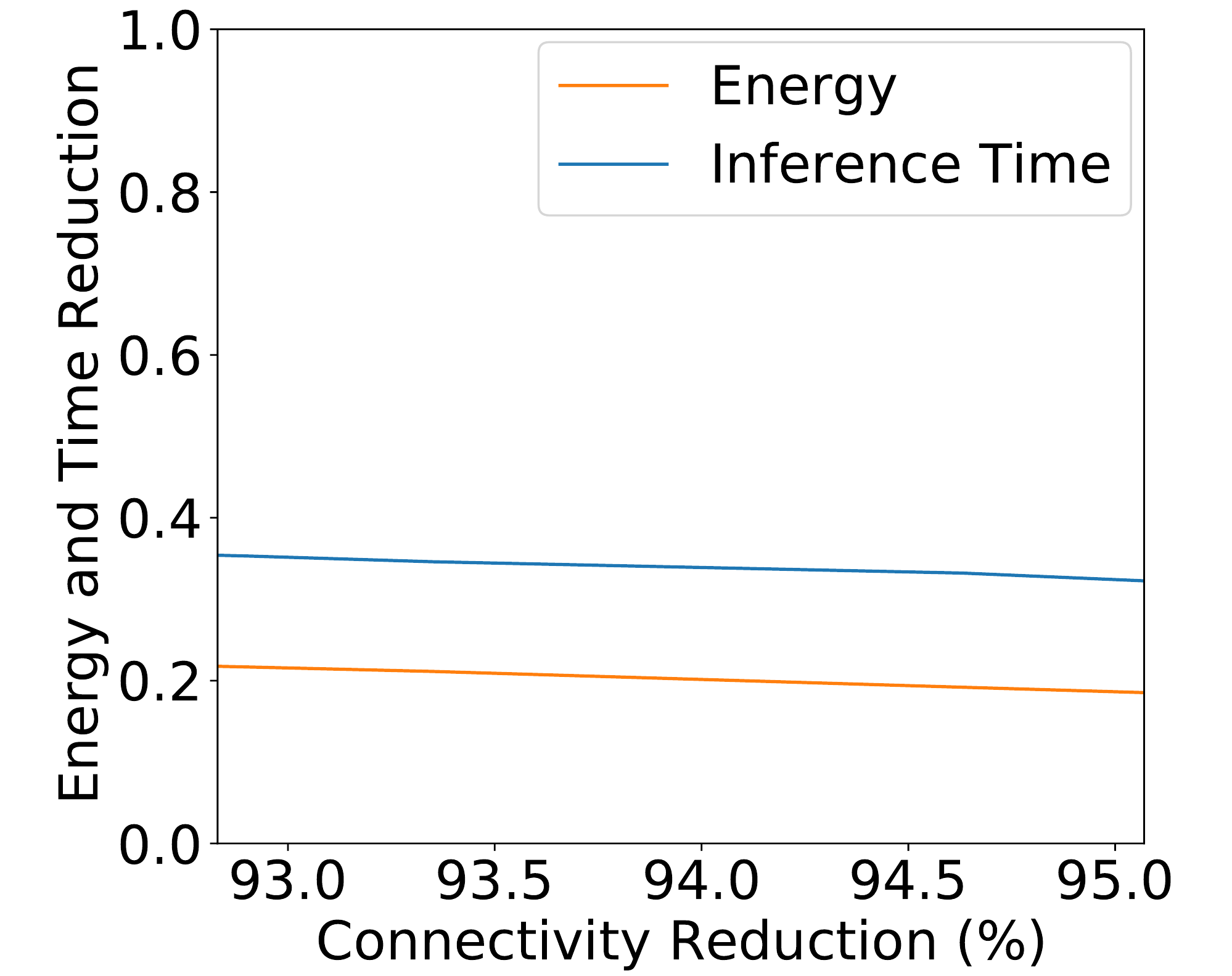}
	\caption{Reduction in energy consumption and response time in the inference, computed as $E_{pruned}/E_{noprune}$ and $t_{pruned}/t_{noprune}$, respectively. $E_{pruned}$ and $E_{noprune}$ are the energy consumption in the inference with and without pruning, $t_{pruned}$ and $t_{noprune}$ are the inference time with and without pruning.}
	\Description{Time speed-up and energy reduction in the inference. The inference time speed-up is 2.825x, 2.832x, 2.89x, 3.011x, and 3.1x and the energy consumption is reduced by 78.24\%, 78.32\%, 78.87\%, 80.81\%, and 81.47\% when the number of connections are reduced by 92.83\%, 92.9\%, 93.35\%, 94.63\%, and 95.07\%, respectively.}
	\label{fig:inference_speedup}
\end{figure}
The network compression during the \ac{STDP}-based learning significantly reduces the response time and energy consumption in the inference stage. As shown in Figure \ref{fig:inference_speedup}, the inference time on our hardware implementation is speeded-up by 2.83-3.1x when 92.83-95\% of the connections in the network are eliminated. In addition, the energy consumption is reduced by 78.24-81.47\% as compared to our baseline without pruning. Meanwhile, the number of spikes generated in the inference is preserved at 95\% compared to the \ac{SNN} trained without the connection pruning. Figure \ref{fig:featuremaps} shows the feature maps of different images in the Caltech 101 dataset in different convolutional layers. The first convolutional layer extracts the edges in the images while the deeper layers observe more abstract features. The feature maps obtained in the \ac{SNN} trained with the connection pruning (Figure \ref{fig:maps_pruned}) are similar to the ones obtained in the \ac{SNN} trained without the connection pruning (Figure \ref{fig:maps_noprune}). This demonstrates that our connection pruning approach does not significantly affect the network behaviours in the inference. Therefore, the classification accuracy was preserved.
\subsection{Comparison with Existing Works}
Prior to our work, there have been approaches that eliminate the weak connections and dormant neurons in the well-trained spiking networks \cite{sen2017approximate, chen2018fast}. However, these works are not applicable during the on-chip learning, when the weight values have not converged. This limits the potential of these works to be applied to the applications that require frequent online learning \cite{sahoo2018online, akusok2019spiking, skatchkovsky2020federated}. On the other hand, the approaches in \cite{rathi2018stdp, shi2019soft, kundu2021spike} and our work eliminate the connections during the on-chip learning. While the works in \cite{shi2019soft, kundu2021spike} focus on the \ac{SNN} with the rate-based coding, our work proposes a connection pruning approach for the \ac{SNN} with the \ac{TTFS} coding, which is more energy-efficient \cite{rullen2001rate}. Note that it is more challenging to eliminate the connections for the \ac{SNN} with the \ac{TTFS} coding, as the spike activities are sparser than the rate-based coding. The connection pruning may cause the number of spikes to drop drastically, causing the shortage of spike activities to trigger the weight updates in the \ac{STDP}-based learning. In addition, our connection pruning approach was evaluated on a deep \ac{SNN}, consisting of three convolutional layers, while the works in \cite{rathi2018stdp, bogdan2018structural, shi2019soft} was applied to the \acp{SNN} consisting of one layer. It is more challenging to eliminate the connections during the \ac{STDP}-based learning in a deep \ac{SNN} as compared to an \ac{SNN} with a single layer. In the deep \ac{SNN}, the errors caused by the overly aggressive pruning in a layer will be carried to the following layers and may eventually increase accuracy loss. 
\section{Conclusions} \label{sec:conclusions}
In this paper, we have proposed a novel connection pruning approach to be applied during the on-chip \ac{STDP}-based learning for \ac{SNN}, in which the spikes are encoded using \ac{TTFS}. Our main contributions are three-fold. First, we proposed a novel connection pruning approach that helps to compress the network and speed-up the on-chip learning time on embedded systems hardware. Our connection pruning approach was evaluated on a deep convolutional \ac{SNN} and achieved the compression of 92.83\% without incurring any loss in the classification accuracy. Second, evaluation on the \ac{FPGA} platform showed that the on-chip learning time was reduced by 2.1x and the inference time was reduced by 2.83x as compared to our baseline with the connection pruning turned off. Moreover, the energy consumption was reduced by 64\% in the on-chip learning and 78.24\% in the inference as compared to the baseline. The hardware implementation of the connection pruning incurs 0.56\% power overhead as compared to our hardware implementation with the pruning units removed. Third, while most of the existing works perform the compression on a well-trained network, our connection pruning approach can be applied during the on-chip \ac{STDP}-based learning. In addition, our approach is applicable to the \ac{SNN} with the \ac{TTFS} coding while most of the existing works focus on the \ac{SNN} with the rate-based coding, which consumes more energy. In the future, our approach can be combined with various device-level optimizations to further reduce the delay, hardware resources, and energy consumption of \ac{SNN} implementations on embedded systems platforms.

\begin{acks}
	This work is supported in part by the NUS Start-up Grant, in part by the Ministry of Education (Singapore) Academic Research Fund (Tier 1), and in part by the A*STAR Programmatic Research Grant (SpOT-LITE).
\end{acks}

\bibliographystyle{ACM-Reference-Format}
\bibliography{references}

\end{document}